\documentclass[journal,twoside,web]{ieeecolor}
\usepackage{tmi}
\usepackage{cite}
\usepackage{amsmath,amssymb,amsfonts}
\usepackage{algorithmic}
\usepackage{graphicx}
\usepackage{textcomp}

\usepackage{multirow}
\usepackage{colortbl}
\usepackage{xcolor}
\usepackage{circledsteps}
\usepackage[bookmarks=true]{hyperref}

\def\BibTeX{{\rm B\kern-.05em{\sc i\kern-.025em b}\kern-.08em
    T\kern-.1667em\lower.7ex\hbox{E}\kern-.125emX}}
\begin{document}
\title{Modeling Annotator Preference and Stochastic Annotation Error for Medical Image Segmentation}

\author{Zehui Liao, Shishuai Hu, Yutong Xie, and Yong Xia, \IEEEmembership{Member, IEEE}
\thanks{This work was supported in part by the National Natural Science Foundation of China under Grants 62171377, in part by the Natural Science Foundation of Ningbo City, China, under Grant 2021J052, in part by the CAAI-Huawei MindSpore Open Fund under Grant CAAIXSJLJJ-2020-005B, and in part by the Innovation Foundation for Doctor Dissertation of Northwestern Polytechnical University.
({\em Z. Liao and S. Hu contributed equally to this work.}
{\em Corresponding author: Y. Xia})}
\thanks{Z. Liao, S. Hu, and Y. Xia are with the National Engineering Laboratory for Integrated Aero-Space-Ground-Ocean Big Data Application Technology, School of Computer Science and Engineering, Northwestern Polytechnical University, Xi'an 710072, China (e-mail: merrical@mail.nwpu.edu.cn; sshu@mail.nwpu.edu.cn; yxia@nwpu.edu.cn). 
Y. Xie is with the Australian Institute for Machine Learning, The University of Adelaide, Adelaide, SA 5000, Australia. (e-mail: yutong.xie678@gmail.com)}
}

\maketitle

\begin{abstract}
Manual annotation of medical images is highly subjective, leading to inevitable and huge annotation biases. Deep learning models may surpass human performance on a variety of tasks, but they may also mimic or amplify these biases.
Although we can have multiple annotators and fuse their annotations to reduce stochastic errors, we cannot use this strategy to handle the bias caused by annotators' preferences. 
In this paper, we highlight the issue of annotator-related biases on medical image segmentation tasks, and propose a Preference-involved Annotation Distribution Learning (PADL) framework to address it from the perspective of disentangling an annotator's preference from stochastic errors using distribution learning so as to produce not only a meta segmentation but also the segmentation possibly made by each annotator. 
Under this framework, a stochastic error modeling (SEM) module estimates the meta segmentation map and average stochastic error map, and a series of human preference modeling (HPM) modules estimate each annotator's segmentation and the corresponding stochastic error. We evaluated our PADL framework on two medical image benchmarks with different imaging modalities, which have been annotated by multiple medical professionals, and achieved promising performance on all five medical image segmentation tasks.
\end{abstract}


\begin{figure}[t]
  \centering
  \includegraphics[width=0.43\textwidth]{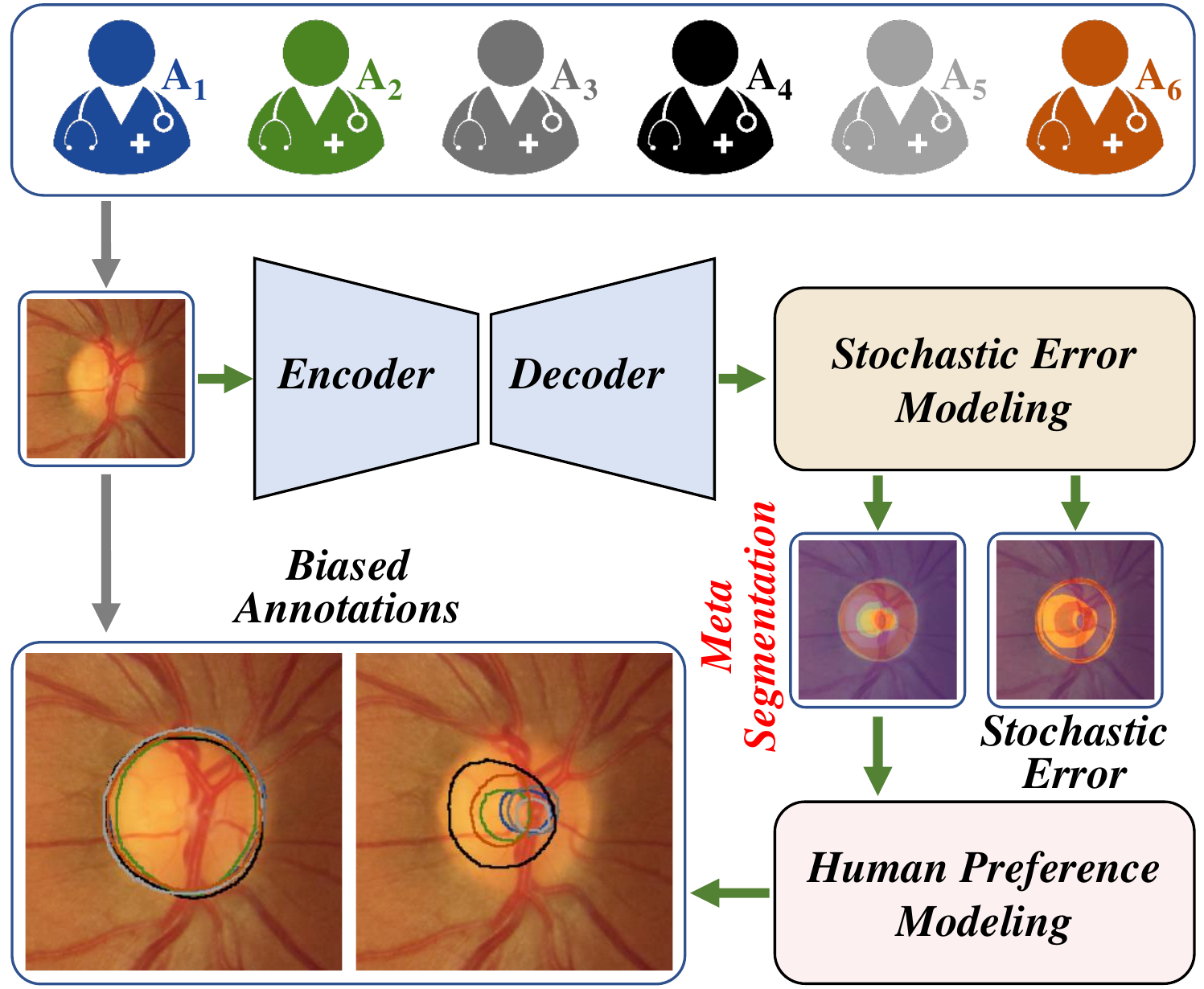}
  \caption{
  Modeling annotation bias in medical image segmentation using the proposed PADL framework.
  Based on their perceptions and expertness, multiple annotators give inconsistent annotations of the same fundus image (indicated by \textcolor[RGB]{127,127,127}{\textbf{gray}} arrows).
  The pipeline of proposed PADL framework is highlighted by \textcolor[RGB]{84,130,53}{\textbf{green}} arrows.
  Under this framework, the meta segmentation map is disentangled from the stochastic error via establishing the annotation distribution, and the annotation made by each annotator (with personal preference) is then predicated based on the meta segmentation map and image features.
  }
  \label{fig:framework}
\end{figure}

\section{Introduction}
\label{sec:intro}
\IEEEPARstart{M}{edical} image segmentation plays a crucial role in delivering effective patient care in diagnostic and treatment practices. 
Among numerous segmentation approaches, deep convolutional neural networks (CNNs) have recently become the de facto standard, providing the state-of-the-art (SOTA) performance on many segmentation tasks~\cite{xie2021unified,Yang2020FDAFD,Zhang2020DoDNetLT,Hu2021DomainAC,Liu2020MSNetMN}. 
Known as data-driven techniques, CNNs require a large scale of accurately annotated images for training, which is indeed impossible to obtain on medical image segmentation tasks.
Besides its tremendous cost, manual annotation of medical images can hardly be \emph{accurate}, since it is highly subjective and relies on observers' perception and expertise~\cite{vincent2021impact,liao2022learning,Fu2014InterraterAI,Taghanaki2018SegmentationfreeDT,Mirikharaji2021DLEMADL}.
For example, three trained observers (two radiologists and one radiotherapist) delineated a lesion of the liver in an abdominal CT image twice with an interval of about one week, resulting in the variation of delineated areas up to 10\% per observer and more than 20\% between observers\footnote{See the \href{https://books.google.com.hk/books?id=PzY6AN2KEoAC&pg=PA216}{figure in this page.}}~\cite{Suetens2017FundamentalsOM}.
The annotator-related bias in ground truths is an `inconvenient truth' in the field of medical image segmentation, whose impact has been rarely discussed.

\begin{figure}[t]
  \centering
  \includegraphics[width=0.45\textwidth]{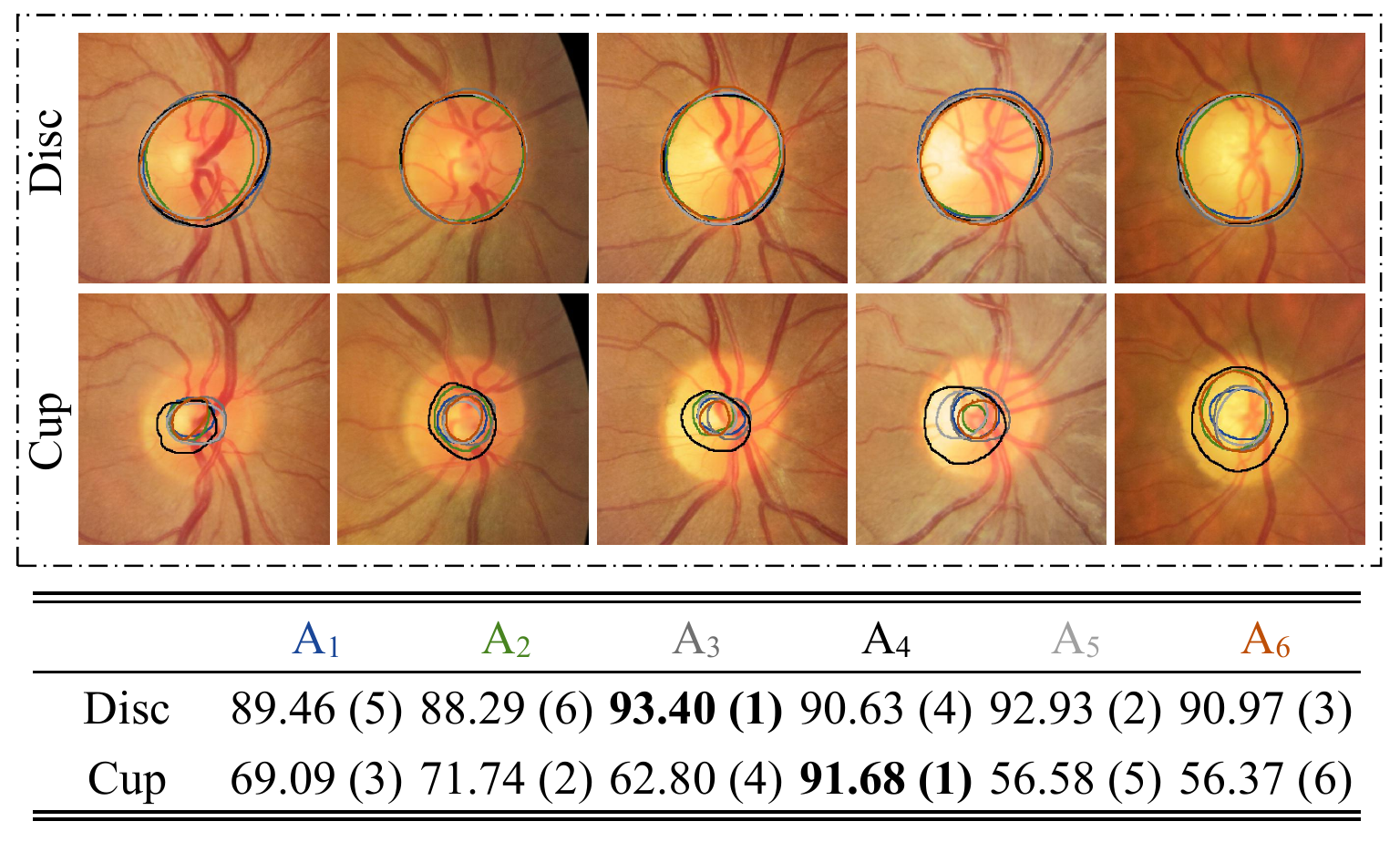}
  \caption{Annotator preference counted from the training set of RIGA. There are five fundus images annotated by six annotators. The IoU between each annotation and the union of six annotations is utilized to quantify the annotator preference. The table shows the average IoU of all training samples. The number in brackets is the rank from highest to lowest in each row.}
  \label{fig:anno_pref}
\end{figure}

To reduce the impact of such annotator-related biases, each training sample can be annotated by multiple medical professionals independently ~\cite{liao2022learning,Fu2014InterraterAI,Taghanaki2018SegmentationfreeDT,Mirikharaji2021DLEMADL} (see figure~\ref{fig:framework}), and a proxy ground truth is generated via majority voting~\cite{Guan2018WhoSW}, label fusion~\cite{Chen2019APLSESSC,Li2020RFNet,Liu2020MSNetMN,Zhang2020Crossdenoise,Zhao2020rvsegbyjointlocallloss,Warfield2004STAPLE}, or label sampling~\cite{Jensen2019Labelsampling}. 
It is worth noting that, in many cases, the variable annotations provided by multiple annotators are all reasonable but with different preferences. For instance, a medical professional who advocates for active treatment usually delineates a slightly larger area of a lesion than the area marked by others.
To illustrate the annotator's preference, we show five fundus images from the RIGA dataset and the annotations of optic disc and optic cup given by six annotators in figure~\ref{fig:anno_pref}. 
The IoU of each annotator's delineation over the union of six annotations is calculated, and the average IoU values over all training samples are listed at the bottom of this figure.
It reveals that the annotator \textbf{$A_3$} prefers to mark much larger optic discs, and the annotator \textbf{$A_4$} and \textbf{$A_2$} prefer to mark larger optic cups.
Using proxy ground truths can somehow diminish the impact of stochastic annotation errors~\cite{Monteiro2020StochasticSN,Mirikharaji2019LearningTS}, but cannot tackle the annotator's preference~\cite{ji2021learning,Guan2018WhoSW,Mirikharaji2019LearningTS,Ribeiro2019HandlingIA,Lampert2016AnES}. Particularly, converting the multiple annotations of each training image into a proxy ground truth not only overlooks the rich information embedded in those annotations, but, more important, may lead the segmentation result to be neither fish nor fowl.
Therefore, instead of reducing the impact of annotation differences, we advocate to disentangle annotators' preference from stochastic annotation errors and characterize both statistically so that a CNN is able to not only produce objective image segmentation, namely meta segmentation, but also mimic each annotator and segment medical images with his or her preference (see figure~\ref{fig:framework}).

To this end, we propose a Preference-involved Annotation Distribution Learning (PADL) framework to address the issue of annotator-related bias in medical image segmentation. 
Under this framework, there are an encoder-decoder backbone, a stochastic error modeling (SEM) module, a series of human preference modeling (HPM) modules, and a series of Gaussian Sampling modules.
The encoder-decoder backbone performs feature extraction.
The SEM module uses image features to estimate the meta segmentation map $\mu$ and average stochastic error map $\sigma$ via annotation distribution approximation. This module also contains an entropy guided attention (EGA) block, where the entropy map of $\mu$ serves as the attention to guide the estimation of $\sigma$. 
In the $r$-th HPM module, a preference estimation block uses the combination of meta segmentation and image features to estimate the $r$-th annotator's segmentation map $\mu_r$, and an EGA block estimates the corresponding stochastic error map $\sigma_r$.
The SEM module and each HPM module is equipped with a Gaussian Sampling module, which samples a probabilistic segmentation map from the Gaussian distribution established by the estimated $\mu$ (or $\mu_r$) and $\sigma$ (or $\sigma_r$).
The loss function is composed of the meta segmentation loss and annotator-specific segmentation loss, each being defined as the binary cross-entropy loss between the sampled segmentation maps and annotations.
We have evaluated the proposed PADL framework on two medical image segmentation benchmarks, which include multiple imaging modalities and five segmentation tasks and are annotated by multiple medical professionals.
To summarise, the contributions of this work are three-fold.
\begin{itemize}
\setlength{\parskip}{0.1em} 
\item[$\bullet$] We highlight the issue of annotator-related biases existed in medical image segmentation tasks, and propose the PADL framework to address it from the perspective of disentangling an annotator's preference from stochastic errors so as to produce not only a meta segmentation but also the segmentation possibly made by each annotator.
\item[$\bullet$] We treat the annotation bias as the combination of an annotator's preference and stochastic errors, and hence design the SEM module and annotator-specific HPM module to characterize each annotator's preference while diminishing the impact of stochastic errors.
\item[$\bullet$] Our PADL framework achieves superior performance against other methods tackling this issue on two medical image segmentation benchmarks (five tasks) with multiple annotators.
\end{itemize}

\begin{figure*}[t]
  \centering
  \includegraphics[width=0.88\textwidth]{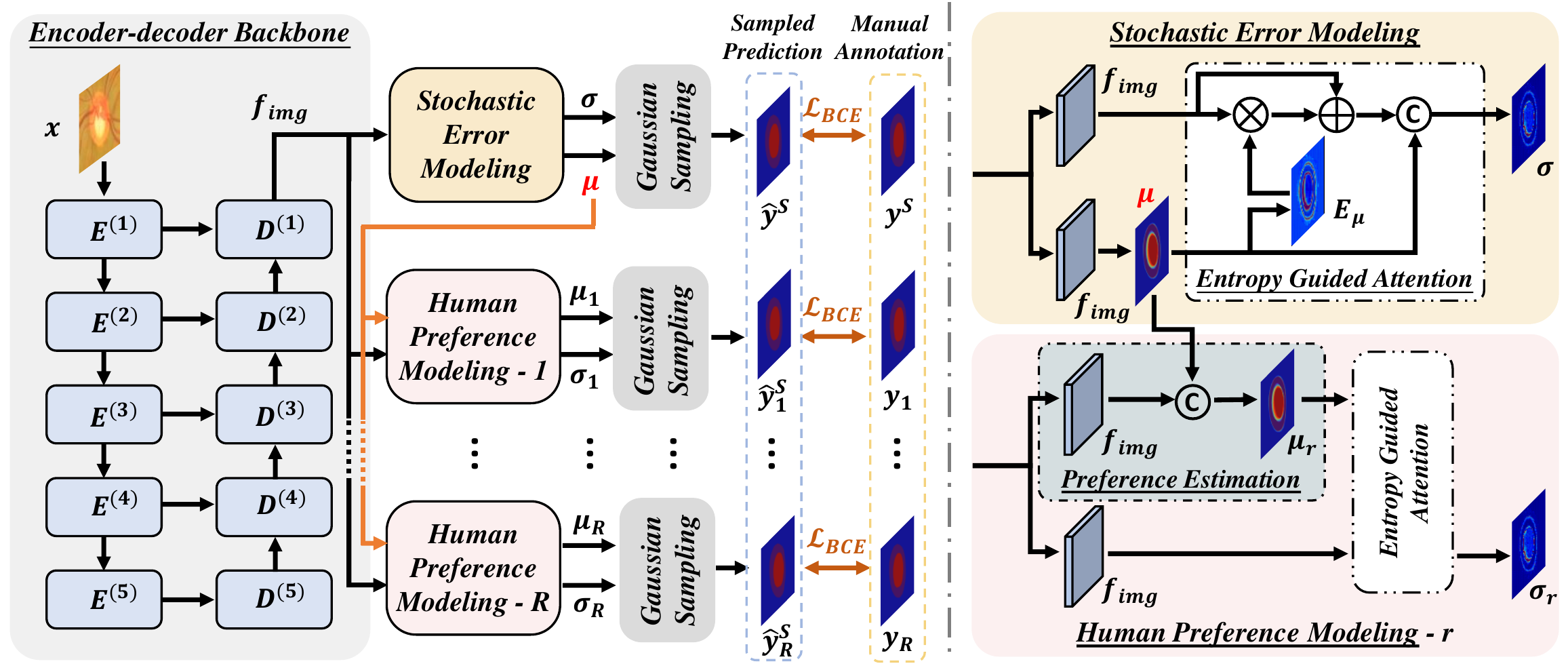}
  \caption{
  Architecture of PADL framework. Our PADL contains an encoder-decoder backbone, a SEM module, $R$ HPM modules, and $R+1$ Gaussian Sampling modules, where $\hat{y}^{S}$ (or $\hat{y}^{S}_{r}$) represents the randomly sampled prediction from $\mathcal{N}\left ( \mu ,\sigma^{2}  \right )$ (or $\mathcal{N}\left ( \mu_{r} ,\sigma_{r}^{2}  \right )$), $y_{r}$ denotes the annotation from $r$-th annotator, $y^{S}$ is the randomly sampled annotation from $R$ annotators, $R$ is the number of annotators, $\mu$ and $\mu_{r}$ are the meta segmentation and the $r$-th annotator's preference-involved segmentation, and $\sigma$ and $\sigma_{r}$ are the average stochastic error and annotator-specific stochastic error.
  The SEM module is shown at top right, and the HPM module is shown at bottom right, where \Circled{c}, $\oplus$, and $\otimes$ denote the concatenation, element-wise addition, and element-wise multiplication, respectively, and $E_{\mu}$ means the entropy of $\mu$. Note that only the $r$-th HPM module is shown due to the space limit.}
  \label{fig:overview}
\end{figure*}

\section{Related Work}
\subsection{Medical Image Segmentation with Multiple Annotators}
A few methods have been proposed to address the issue of annotator-related bias in medical image segmentation, which can be roughly grouped into two categories.
\subsubsection{Annotator Decision Fusion Methods} 
This kind of method usually uses multiple convolutional heads to model annotators, respectively, and calculate the weighted sum of multi-head outputs as the segmentation result~\cite{Guan2018WhoSW}.
Based on the annotator re-weighting framework, Xiao et al.~\cite{xiao2021pathological} further propose a Gaussian attention focal loss that makes the model pay more attention to essential regions. 
Mirikharaji et al.~\cite{Mirikharaji2019LearningTS} take the professional level of annotators into account and propose a sample re-weighting approach to assign higher importance to samples annotated by professionals in the loss function.
Recently, the uncertainty map that measures the disagreements among all annotators has been adopted as the spatial attention to calibrate the segmentation result~\cite{ji2021learning}.
Despite their advantages, these methods are sensitive to stochastic annotation errors~\cite{jungo2018effect}.
\subsubsection{Annotator Bias Disentangling Methods} 
With the advance of image classification using noisy labels~\cite{Tanno2019LearningFN}, a typical annotator bias disentangling method that uses two coupled CNNs has been proposed to disengage annotator bias from these inconsistent annotations~\cite{Zhang2020DisentanglingHE,Zhang2020Learntoseg}.
The segmentation CNN estimates the label distribution, and the annotation CNN models the human bias using a confusion matrix.
Although achieving promising performance under elaborated hyper parameters, this method cannot estimate the expected label distribution effectively due to the undesired constraint of the confusion matrix.

By contrast, our PADL framework establishes the annotation distribution to disentangle the stochastic errors from manual annotations to reduce its negative impact. Meanwhile, annotators' preference is modeled as a series of annotation transformations using a stack of convolution layers.

\subsection{Annotation Distribution Learning}
The research on annotation distribution learning can be traced back to the seminal work of probabilistic U-Net (PU-Net)~\cite{Kohl2018PUNet}, which combines a U-Net~\cite{Ronneberger2015UNetCN} with a conditional variational autoencoder (VAE) to form a generative segmentation model that is able to produce an unlimited number of plausible segmentation results.
Thanks to the development of VAE and reparameterization~\cite{kingma2013auto}, annotation distribution learning has recently been increasingly studied~\cite{Kohl2018PUNet,Hu2019SupervisedUQ,Kohl2019AHP,baumgartner2019phiseg,Kassapis2020CARSSS,liu2022variational}, Hierarchical Probabilistic U-Net (HPU-Net)~\cite{Kohl2019AHP} was constructed to improve the segmentation reconstruction fidelity by introducing a hierarchical latent space decomposition.
An adversarial refinement method~\cite{Kassapis2020CARSSS} was proposed for stochastic semantic segmentation, which employs a generative adversarial network~\cite{goodfellow2014generative} to calibrate the predicted distribution over semantic maps.
These methods model the distribution at the feature level for capturing such ambiguities in manual annotations. 
By contrast, our PADL employs annotation distribution learning to directly approximate the distribution from the biased annotations at the output level so that the impact of stochastic errors can be diminished.
Furthermore, we model human preference as annotation transformations so as to further reconstruct each annotator's segmentation.

\section{Method}
\subsection{Problem Definition and Method Overview}
Let a set of medical images annotated by $R$ annotators be denoted by $D=\left \{ x_{i}, y_{i1}, y_{i2}, \cdots, y_{iR} \right \}_{i=1}^{N}$, where $x_{i} \in \mathbb{R}^{C\times H\times W}$ represents the $i$-th image with $C$ channels and a size of $H\times W$, and $y_{ir} \in \left \{ 0,1 \right \} ^{K\times H \times W}$ is the annotation with $K$ classes given by the $r$-th annotator. 
Our goal is to train a segmentation model on the training set $D_{tr}$ so that the model can generate meta segmentation and mimic each annotator, and perform segmentation on the testing set $D_{ts}$ with his/her preference. 

The proposed PADL framework consists of an encoder-decoder backbone, a SEM module, $R$ HPM modules, and $R+1$ Gaussian Sampling modules (see Fig.~\ref{fig:overview}).
For each input image $x$, the backbone extracts its feature map $f_{img}$.
Based on $f_{img}$, the SEM module estimates a meta segmentation map $\mu$, which approximates the mean voting of $R$ annotations, and an average stochastic error map $\sigma$.
Based on $\mu$ and $f_{img}$, a HPM module can estimate the segmentation map and stochastic error map of each annotator, denoted by $\mu_r$ and $\sigma_r$, respectively.
With the established distributions, the Gaussian Sampling module can sample a segmentation result, which is compared with the corresponding annotation to generate the supervisory signal.
We now delve into the details of our PADL framework.

\subsection{Backbone}
The backbone network has a U-like encoder-decoder structure~\cite{Ronneberger2015UNetCN}, where we adopt the ResNet34~\cite{He2016DeepRL} pretrained on the ImageNet dataset~\cite{Deng2009Imagenet} as the encoder. 
To adapt it to our tasks, we replace the last average pooling layer and fully connection layer in ResNet34 with a ReLU layer. 
Skip connections are from the convolutional block and first three residual blocks in the encoder to the corresponding locations in the decoder~\cite{ji2021learning}.

Symmetrically, the decoder is composed of five blocks, which upsample the feature map gradually to restore its resolution. 
In each of the first four blocks, the feature map is upsampled by a transposed convolutional layer with a stride of 2, processed by a convolutional layer, concatenated with the feature map from the encoder, and fed to a ReLU layer and a batch normalization layer.
The last decoder block only upsamples the image features using a transposed convolutional layer with a stride of 2.
As a result, the decoder produces a 32-channel feature map $f_{img}$ for each input image, shown as follows
\begin{equation}
    f_{img} = \mathrm {F}_{D} \left ( \mathrm {F}_{E} \left ( x;\theta _{E}  \right ) ;\theta _{D}  \right ), 
\end{equation}
where $\theta_{E}$ and $\theta_{D}$ represent the parameters in the encoder $F_E$ and decoder $F_D$, respectively.

\subsection{Stochastic Error Modeling}
We assume the annotation distribution to be Gaussian.
To approximate this distribution, we need estimate the mean $\mu \in \mathbb{R}^{K \times H\times W}$ and standard deviation $\sigma \in \mathbb{R}^{K \times H\times W}$ on a pixel-by-pixel basis.
When using all annotations, $\mu$ is an estimation of the mean voting of $R$ annotations, and hence is called the meta segmentation map. The average stochastic error existed in all annotations is measured by $\sigma$.
We use a $\mu$ head $\mathrm{F_{\mu}}$, which is a $1\times1$ convolutional layer, to estimate $\mu$ as follows
\begin{equation}
    \mu = \mathrm{F_{\mu}} \left ( f_{img};\theta_{\mu} \right ),
\end{equation}
where $\theta _{\mu}$ represents the parameters in $\mathrm{F_{\mu}}$.

We use a $\sigma$ head $\mathrm{F_{\sigma}}$, which is composed of a $1\times1$ convolutional layer, a batch normalization layer and a ReLU layer, to produce the $\sigma$ features $f_{\sigma}$ as follows
\begin{equation}
    f_{\sigma} = \mathrm{F_{\sigma}} \left ( f_{img};\theta_{\sigma} \right ), 
\end{equation}
where $\theta_{\sigma}$ represents the parameters in $\mathrm{F_{\sigma}}$.

\subsubsection{Entropy Guided Attention Block}
The stochastic annotation errors always occur in ambiguous regions, where the entropy of manual annotations is high~\cite{Kassapis2020CARSSS}. Therefore, we design an EGA block to process and calibrate $f_{\sigma}$.
Given the meta segmentation map $\mu$, its entropy map $E_{\mu}$ can be calculated as
\begin{equation}
    E_{\mu} = -\mu \times \log_2(\mu) - (1-\mu) \times \log_2(1-\mu).
\end{equation}
We use the entropy map $E_{\mu}$ as the spatial attention~\cite{woo2018cbam} to highlight ambiguous regions. Thus, $f_{\sigma}$ can be calibrated as
\begin{equation}
    \widetilde{f}_{\sigma} = f_{\sigma} \times (1 + E_{\mu}).
\end{equation}
Since the stochastic error $\sigma$ is highly related to the meta segmentation $\mu$, we concatenate $\mu$ with $\widetilde{f}_{\sigma}$ to predict $\sigma$, shown as follows
\begin{equation}
    \sigma = \mathrm{F_{\sigma}^o}((\mu\  \Circled{c}\  \widetilde{f}_{\sigma});\theta_{\sigma}^o),
\end{equation}
where \Circled{c} represents concatenation, $\mathrm{F_{\sigma}^o}$ is a $1\times1$ convolutional layer, and $\theta_{\sigma}^o$ denotes the parameters in $\mathrm{F_{\sigma}^o}$.

\subsection{Human Preference Modeling}
We design $R$ HPM modules to characterize the preference of $R$ annotators, respectively. Each HPM module contains a preference estimation block and an EGA block. The former estimates the annotator-specific segmentation map $\mu_r$, and the latter estimates the annotator-specific stochastic error $\sigma_r$.

\subsubsection{Preference Estimation}
Due to the annotator's preference, some delineated areas are larger than others and some are smaller, which can be mimicked by using morphological dilation or erosion.
Inspired by this, the preference estimation block is implemented as a stack of convolutional layers.
Here, we use two $3\times3$ convolutional layers with a padding of 1 due to the trade-off between optimization complexity and preference modeling capacity (see ablation study in Table~\ref{tab:Ablation-hp-capacity}).
We first use a $1\times1$ convolutional layer with batch normalization and ReLU activation, denoted by $\mathrm{F_{p}^r}$, to reduce the channel of feature map $f_{img}$ to $K$.
Then, we concatenate the channel-reduced feature map with meta segmentation $\mu$, and feed the concatenation to the preference estimation block $F_{\mu}^r$ for the estimation of preference-involved segmentation map $\mu_r$, shown as follows
\begin{equation}
    \mu_r = \mathrm{F_{\mu}^r}((\mu\  \Circled{c}\  \mathrm{F_{p}^r}(f_{img}; \theta_{p}^r)); \theta_{\mu}^r),
\label{eq:pref_modeling_architecture}
\end{equation}
where $\theta_{p}^r$ and $\theta_{\mu}^r$ represents the parameters in $\mathrm{F_{p}^r}$ and $\mathrm{F_{\mu}^r}$, respectively.

Since the stochastic error occurs in each annotator's delineations, we also use the EGA block to estimate the annotator-specific stochastic error $\sigma_r$.

\subsection{Gaussian Sampling}
With the Gaussian assumption and estimated $\mu$ and $\sigma$, the probabilistic segmentation map $\hat{y}^{s}$ can be sampled from the distribution $\mathcal{N}\left ( \mu ,\sigma^{2}  \right )$ on a pixel-by-pixel basis.
Then, we use the sigmoid function to map those sample values to $(0,1)$ and normalize them to be probabilities.
The probabilistic meta segmentation map $\hat{y}$ is calculated by applying the sigmoid function to the estimated $\mu$ followed by normalization.

Similarly, the annotator-specific segmentation prediction maps $\{\hat{y}_r\}_{r=1}^R$ and corresponding sampled prediction maps $\{\hat{y}_r^{s}\}_{r=1}^R$ can be obtained via applying sigmoid function to the preference-involved segmentation map $\{\mu_r\}_{r=1}^R$ and the probabilistic segmentation map sampled from the established distribution $\{\mathcal{N}\left ( \mu_r ,\sigma_r^{2}  \right )\}_{r=1}^R$.

\subsection{Loss and Inference}
\subsubsection{Loss}
The loss of our PADL framework consists of two parts: the meta segmentation loss $\mathcal{L}_{meta}$ and annotator-specific segmentation loss $\mathcal{L}_{pref}$, shown as follows
\begin{equation}
\begin{aligned}
    \mathcal{L} &= \mathcal{L}_{meta}(y^s, \hat{y}^s) + \sum_{r=1}^R\mathcal{L}_{pref}(y_r, \hat{y}_r^s),
\end{aligned}
\end{equation}
where $y^s$ is randomly selected from all annotations per image, $y_r$ is the delineation given by annotator $A_r$, and the meta segmentation loss is the following binary cross-entropy loss
\begin{equation}
    \mathcal{L}_{meta}(y^s, \hat{y}^s) = -y^s\times\log{\hat{y}^s} - (1-y^s)\times\log{(1-\hat{y}^s)}.
\end{equation}
The annotator-specific segmentation loss $\mathcal{L}_{pref}$ is also the binary cross-entropy loss and is calculated in a similar way.

\subsubsection{Inference} 
During inference, the estimated $\mu$ in the SEM module is adopted as meta segmentation, and the probability map $\hat{y}$ is used to evaluate with mean voting annotation $\bar{y}$.
The approximated $\mu_r$ in each HPM module is viewed as the predicted annotator-specific segmentation map involving human preference, and its corresponding probability map $\hat{y}_r$ is utilized to measure the similarity with delineation $y_r$ from annotator $A_r$.

\begin{table*}[h]
\small
\renewcommand\arraystretch{1.05}
\setlength\tabcolsep{1.5pt}
\centering
\caption{Performance of our PADL framework, 11 competing methods, and two variants of PADL in optic disc and optic cup segmentation on RIGA.
The soft Dice ($\mathcal{D}_{disc}^{s}$ (\%), $\mathcal{D}_{cup}^{s}$ (\%)) are used as the performance metric.
The predictions of each model are evaluated against each annotator's delineations and the mean voting annotation, and the average performance over six annotations is also given.
Except for the results of Multi-Net ($M_1\sim M_6$) and two PADL variants, top three results in each column are highlighted in \textcolor{red}{red}, \textcolor{blue}{blue} and \textcolor[RGB]{0,100,0}{green}, respectively.
In each column, the best result among $M_1\sim M_6$ is highlighted with \underline{underline}.
The cells in \textcolor{gray}{\textbf{gray}} represent the U-Nets trained and tested using the annotations from the same annotator.
}
\begin{tabular}{l|c|c|c|c|c|c|c|c}
\hline \hline
Models & $\rm A_1$  & $\rm A_2$    & $\rm A_3$    & $\rm A_4$    & $\rm A_5$    & $\rm A_6$  & \textbf{Average}    & \textbf{Mean Voting}  \\ \hline \hline
$\rm M_1$ & \cellcolor{lightgray}(\underline{96.16}, 84.29) & (95.08, 80.79) & (95.57, 79.82) & (96.29, 78.91) & (95.91, 80.49) & (96.47, 76.57) & (95.91, 80.15)  & (96.27, 80.56) \\ \hline
$\rm M_2$ & (95.72, \underline{84.71}) & \cellcolor{lightgray}(\underline{95.50}, \underline{84.20}) & (95.52, 79.87) & (96.13, 81.16) & (96.13, 80.91) & (96.27, 77.93) & (95.88, \underline{81.46})  & (96.30, \underline{82.03}) \\ \hline
$\rm M_3$ & (95.10, 82.76) & (94.50, 79.69) & \cellcolor{lightgray}(\underline{96.53}, \underline{83.10}) & (96.20, 78.39) & (96.28, 81.47) & (95.92, 76.73) & (95.76, 80.36)  & (95.89, 80.90) \\ \hline
$\rm M_4$ & (95.92, 81.46) & (95.30, 82.16) & (96.18, 78.53) & \cellcolor{lightgray}(\underline{96.79}, \underline{87.90}) & (\underline{96.84}, 74.47) & (96.43, 70.57) & (\underline{96.24}, 79.18)  & (96.44, 78.94) \\ \hline
$\rm M_5$ & (95.27, 82.93) & (94.83, 79.99) & (96.27, 81.62) & (96.39, 75.94) & \cellcolor{lightgray}(96.69, \underline{83.15}) & (95.91, 77.64) & (95.89, 80.21)  & (96.08, 81.02) \\ \hline
$\rm M_6$ & (95.92, 80.94) & (95.31, 78.48) & (96.23, 78.14) & (96.56, 73.62) & (96.45, 81.64) & \cellcolor{lightgray}(\underline{96.90}, \underline{80.45}) & (96.22, 78.88)  & (\underline{96.55}, 80.23) \\ \hline
\hline
MH-UNet~\cite{Guan2018WhoSW} & (\textcolor{blue}{96.36}, \textcolor[RGB]{0,100,0}{83.49}) & (\textcolor[RGB]{0,100,0}{95.32}, \textcolor{blue}{81.84}) & (\textcolor{red}{96.75}, 77.20) & (\textcolor{red}{97.01}, \textcolor{blue}{88.21}) & (\textcolor{red}{97.15}, 78.95) & (\textcolor{red}{97.22}, 75.85) & (\textcolor{red}{96.64}, \textcolor[RGB]{0,100,0}{80.92})  & (97.41, 85.21) \\ \hline
MV-UNet~\cite{Ronneberger2015UNetCN} & (95.12, 76.65) & (94.57, 78.12) & (95.55, 77.74) & (95.79, 76.31) & (95.87, 78.67) & (95.68, 74.80) & (95.43, 77.05)  & (97.42, \textcolor[RGB]{0,100,0}{86.11}) \\ \hline
LS-UNet~\cite{Jensen2019Labelsampling} & (95.43, 75.66) & (94.82, 74.56) & (95.57, 73.52) & (95.96, 72.30) & (95.90, 75.72) & (95.93, 72.85) & (95.60, 74.10)  & (\textcolor{blue}{97.58}, 82.68) \\ \hline
\hline
MR-Net~\cite{ji2021learning} & (95.35, 81.77) & (94.81, 81.18) & (95.80, \textcolor[RGB]{0,100,0}{79.23}) & (95.96, 84.46) & (95.90, \textcolor[RGB]{0,100,0}{79.04}) & (95.76, \textcolor[RGB]{0,100,0}{76.20}) & (95.60, 80.31)  & (\textcolor[RGB]{0,100,0}{97.55}, \textcolor{blue}{87.20}) \\ \hline
CM-Net~\cite{Zhang2020DisentanglingHE} & (\textcolor[RGB]{0,100,0}{96.29}, \textcolor{blue}{84.59}) & (\textcolor{blue}{95.46}, \textcolor[RGB]{0,100,0}{81.44}) & (\textcolor[RGB]{0,100,0}{96.60}, \textcolor{blue}{81.84}) & (\textcolor{blue}{96.90}, \textcolor[RGB]{0,100,0}{87.52}) & (\textcolor{blue}{96.86}, \textcolor{blue}{82.39}) & (\textcolor{blue}{96.93}, \textcolor{blue}{78.82}) & (\textcolor[RGB]{0,100,0}{96.51}, \textcolor{blue}{82.77})  & (96.64, 81.96) \\ \hline
\hline 
Ours & (\textcolor{red}{96.40}, \textcolor{red}{85.22}) & (\textcolor{red}{95.60}, \textcolor{red}{85.15}) & (\textcolor{blue}{96.64}, \textcolor{red}{82.76}) & (\textcolor[RGB]{0,100,0}{96.82}, \textcolor{red}{88.79}) & (\textcolor[RGB]{0,100,0}{96.78}, \textcolor{red}{83.45}) & (\textcolor[RGB]{0,100,0}{96.87}, \textcolor{red}{79.72}) & (\textcolor{blue}{96.52}, \textcolor{red}{84.18})  & (\textcolor{red}{97.65}, \textcolor{red}{87.75}) \\ \hline \hline
Ours w/o SEM & (96.49, 84.87) & (95.69, 83.13) & (96.42, 83.70) & (96.93, 88.73) & (96.64, 81.99) & (96.77, 79.50) & (96.49, 83.65)  & (96.42, 85.37) \\ \hline
Ours w/o HPM & (95.70, 81.62) & (95.17, 79.95) & (96.10, 79.38) & (96.43, 78.26) & (96.37, 80.01) & (96.27, 76.21) & (96.06, 79.24)  & (97.71, 87.56) \\ \hline
\hline
\end{tabular}
\label{tab:RIGA-comparision}
\end{table*}

\section{Experiments}
\subsection{Datasets}
\subsubsection{RIGA} The RIGA benchmark~\cite{almazroa2017agreement} is collected for the evaluation of optic cup/disc segmentation algorithms. It contains 750 color fundus images from three sources, including 460 images from MESSIDOR, 195 images from BinRushed, and 95 images from Magrabia. 
Six ophthalmologists from different eye centers labeled the optic cup/disc contours manually in each image. 
We followed the data split scheme used in~\cite{yu2019robust,ji2021learning}, using 655 samples from BinRushed and MESSIDOR for training and 95 samples from Magrabia for test.

\subsubsection{QUBIQ} The QUBIQ dataset~\cite{QUBIQ2021zenodo}
is a dataset specifically collected to evaluate inter-annotator variability.
It contains four subsets, including 39 MRI cases (34 for training and 5 for test) with seven annotators for brain growth segmentation, 32 MRI cases (28 for training and 4 for test) with three annotators for brain tumor segmentation, 55 MRI cases (48 for training and 7 for test) with six annotators for prostate and center zone segmentation, and 24 CT cases (20 for training and 4 for test) with three annotators for kidney segmentation. The dataset split was provided by~\cite{ji2021learning}.

\subsection{Experimental Setup}
\subsubsection{Implementation Details}
For each segmentation task, all images were normalized via subtracting the mean and dividing by the standard deviation on a pixel-by-pixel basis. The mean and standard deviation were counted on training cases.
For a fair comparison, we followed the settings in~\cite{ji2021learning}, $i.e.$, setting the mini-batch size to 8 and resizing the input image to $256\times256$ for all tasks.
The images in the prostate subset in QUBIQ are relatively large, and hence were center cropped to $640 \times 640$ before resizing to $256 \times 256$.
The Adam optimizer~\cite{Kingma2015AdamAM} with a initial learning rate of $lr_{0}=1e-4$ was adopted as the optimizer. 
The learning rate was decayed according to the polynomial policy $lr = lr_{0} \times \left ( 1-t/T  \right ) ^{0.9}$, where $t$ is the current epoch and $T$ is the maximum epoch.
The maximum epoch $T$ was set to 60 for optic cup and optic disc segmentation and 1000 for QUBIQ segmentation tasks.
Other parameters in Adam were set as default.
All experiments were performed on a workstation with one NVIDIA RTX 2080Ti GPU and implemented under the PyTorch framework.

\subsubsection{Evaluation Metrics} Since each test sample has multiple annotations, we adopted the Soft Dice ($\mathcal{D}^s$) and Soft IoU ($IoU^s$) as performance metrics.
At each threshold level, the threshold is applied to the predicted probability map and mean voting of annotations to generate hard Dice/IoU.
Soft Dice/IoU is calculated via averaging the hard Dice/IoU values obtained at multiple threshold levels, $i.e.$, (0.1, 0.3, 0.5, 0.7, 0.9) for this study.
Based on Soft Dice, there are two performance metrics, namely `Average' and `Mean Voting'.
`Mean Voting' is the Soft Dice between the predicted meta segmentation and the mean voting annotations. Higher `Mean Voting' represents better performance on modeling meta segmentation.
The annotator-specific predictions are evaluated against each annotator’s delineations, and the average Soft Dice is denoted as `Average'.  Higher `Average' represents better performance on miming each annotator.

\subsection{Comparison Results}
\subsubsection{Experiments on RIGA dataset}
On the RIGA dataset, we compared our PADL to (1) the baseline `Multi-Net' setting, under which six U-Nets (denoted by $M_1\sim M_6$) were trained with the annotations provided by annotator $A_1\sim A_6$, respectively;
(2) MH-UNet: a U-Net variant with multiple segmentation heads, each accounting for imitating the annotations from a specific annotator~\cite{Guan2018WhoSW};
(3) MV-UNet: a U-Net trained with the mean voting of annotations~\cite{Ronneberger2015UNetCN};
(4) LS-UNet: a U-Net trained with randomly selected annotation from the candidate annotations of each sample~\cite{Jensen2019Labelsampling};
(5) MR-Net: an annotator decision fusion method that uses an attention module to model the multi-rater agreement~\cite{ji2021learning}; and
(6) CM-Net: an annotator bias disentangling method that uses a confusion matrix to model human errors~\cite{Zhang2020DisentanglingHE}.

The soft Dice of optic disc $\mathcal{D}_{disc}^s$ and optic cup $\mathcal{D}_{cup}^s$ obtained by our model and completing methods were listed in Table~\ref{tab:RIGA-comparision}.
As expected, the U-Net trained using the annotations from annotator $A_{r}$ always achieves higher performance when evaluated against $A_{r}$'s delineations, except that $M_1$ slightly underperforms $M_2$ in $\mathcal{D}_{cup}^{s}$ and $M_5$ slightly underperforms $M_4$ in $\mathcal{D}_{disc}^{s}$.
Also, the mean voting-based proxy ground truth method MV-UNet outperforms the multi-network ($M_1\sim M_6$) and multi-head (MH-UNet) methods when evaluated against the mean voting of annotations, but underperforms both when evaluate against each annotator's delineations.
A similar conclusion can be drawn for the random annotation-based proxy ground truth method LS-UNet, except for its surprisingly low performance in optic cup segmentation. 

Meanwhile, since the annotator-related bias is considered, both MR-Net and CM-Net perform well no matter being evaluated against the mean voting or each individual annotation. Between them, MR-Net outperforms CM-Net when evaluated against the mean voting, but is inferior to CM-Net in reconstructing each annotator's delineations. It may attribute to the fact that the annotator decision fusion strategy can hardly preserve each annotator's preference.
Among 12 competing methods (two variants of PADL not counted in), the proposed PADL achieves the second highest average $\mathcal{D}_{disc}^s$ and highest average $\mathcal{D}_{cup}^s$ when evaluated against each annotator's delineations and achieves the highest soft Dice on both segmentation tasks when evaluated against the mean voting.
It should be noted that, comparing to the annotator bias disentangling method CM-Net, our PADL is able to not only reconstruct each annotator's delineations slightly better but also perform substantially more accurate meta segmentation to approximate the mean voting of annotations.

\begin{figure}[t]
  \centering
  \includegraphics[width=0.48\textwidth]{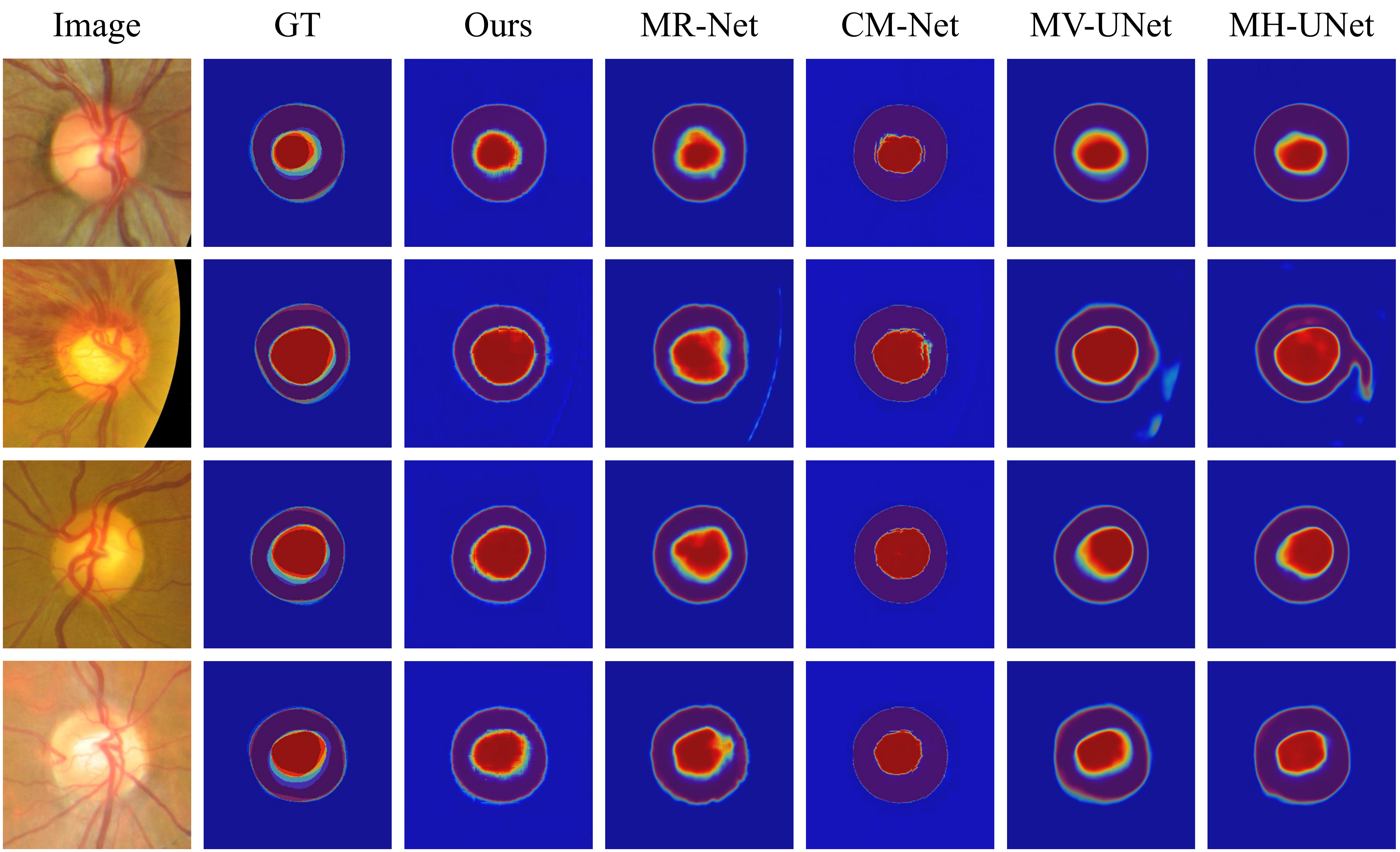}
  \caption{
  Visualization of probabilistic segmentation maps predicted by our PADL, MR-Net, CM-Net, MV-UNet, and MH-UNet for four cases from the RIGA dataset, together with the original image and ground truth (GT).
  }
  \label{fig:visual_RIGA}
\end{figure}

We visualized the probabilistic segmentation maps predicted by our PADL and other four competing methods, $i.e.$, MR-Net, CM-Net, MV-UNet and MH-UNet, and the mean voting annotation maps of four cases from the RIGA dataset in figure~\ref{fig:visual_RIGA}.
It shows that our PADL can produce more accurate probability maps, confirming that our PADL can model the consensus among all annotators via disentangling stochastic errors from the meta segmentation.

\subsubsection{Experiments on QUBIQ dataset}

\begin{table}[t]
\small
\centering
\renewcommand\arraystretch{1.05}
\setlength\tabcolsep{2.5pt}
\caption{
Soft Dice (\%) of our PADL framework and five competing methods obtained on five segmentation tasks using the QUBIQ dataset. 
The ground truth is the majority voting with average weight of annotations from multiple annotators, and the `Mean Voting' of each task is calculated.
}
\begin{tabular}{l|c|c|c|c|c}
\hline \hline
Methods & $\mathcal{D}_{kidney}^{s}$         & $\mathcal{D}_{brain}^{s}$          & $\mathcal{D}_{tumor}^{s}$          & $\mathcal{D}_{pros1}^{s}$      & $\mathcal{D}_{pros2}^{s}$ \\ \hline \hline
MV-UNet~\cite{Ronneberger2015UNetCN} & 70.65          & 81.77          & 84.03          & 85.18          & 68.39        \\ 
LS-UNet~\cite{Jensen2019Labelsampling} & 72.31          & 82.79          & 85.85          & 86.23          & 69.05      \\ 
MH-UNet~\cite{Guan2018WhoSW} & 73.44          & 83.54          & 86.74          & 87.03          & 75.61                \\ 
MR-Net~\cite{ji2021learning}   & 74.97          & 84.31          & 88.40          & 87.27          & 76.01               \\ 
CM-Net~\cite{Zhang2020DisentanglingHE} & 76.01         & 84.75          & 87.37           & 88.73          & 77.39                \\ \hline
Ours    & \textbf{80.34} & \textbf{85.86} & \textbf{89.25} & \textbf{93.30} & \textbf{80.67} \\ \hline
\hline
\end{tabular}
\label{tab:QUBIQ-results}
\end{table}

\begin{table}[t]
\small
\centering
\renewcommand\arraystretch{1.05}
\setlength\tabcolsep{2.5pt}
\caption{
Soft Dice (\%) of CM-Net and our PADL on five tasks of the QUBIQ dataset.
The predictions are evaluated against each annotator's delineations, and the average soft Dice (`Average') over all annotators is given.
}
\begin{tabular}{l|c|c|c|c|c}
\hline \hline
Methods & $\mathcal{D}_{kidney}^{s}$ & $\mathcal{D}_{brain}^{s}$ & $\mathcal{D}_{tumor}^{s}$ & $\mathcal{D}_{pros1}^{s}$ & $\mathcal{D}_{pros2}^{s}$ \\ \hline \hline
CM-Net~\cite{Zhang2020DisentanglingHE}        & 80.63 & 83.16 & 87.17 & 87.42 & 73.45 \\ \hline
Ours         & \textbf{82.26} & \textbf{83.19} & \textbf{87.58} & \textbf{92.26} & \textbf{75.16} \\ \hline \hline
\end{tabular}
\label{tab:QUBIQ_comparison}
\end{table}

We conducted five CT or MRI image segmentation tasks using the QUBIQ dataset.
First, we compared the meta segmentation modeling capability of the proposed PADL with other five methods, including:
(1) three commonly used strategies that consider the annotator-related bias,
$i.e.$, MV-UNet~\cite{Ronneberger2015UNetCN}, LS-UNet~\cite{Jensen2019Labelsampling}, and MH-UNet~\cite{Guan2018WhoSW},
(2) two recent methods that attempt to address this issue, $i.e.$, MR-Net~\cite{ji2021learning} and CM-Net~\cite{Zhang2020DisentanglingHE}.
Since CM-Net uses a different backbone, we replaced its backbone with the one used by other methods~\cite{ji2021learning} via revising the publicly released code. We also optimized the hyper-parameters for CM-Net and adopted the same experimental settings ~\cite{ji2021learning} for a fair comparison.
Table~\ref{tab:QUBIQ-results} shows the soft Dice of our PADL and five competing methods obtained by evaluating against the majority voting of annotations. Note that the soft Dice of four competing methods (except for CM-Net) were adopted from ~\cite{ji2021learning}.
It reveals that the soft Dice values of three commonly used strategies are relatively lower than those of others, suggesting that both the proxy ground truth strategy and multi-head strategy suffer from limited performance.
By contrast, our PADL achieves the highest soft Dice on all tasks, beating the proxy ground truth strategy, multi-head strategy, and two recent methods.
The results displayed in this table are consistent with those obtained on the RIGA dataset, confirming the superior ability of our PADL to address the stochastic annotation errors and generate the meta segmentation.

Second, we also compared the annotator's delineations reconstructing ability of the proposed PADL and the best competitor, $i.e.$, CM-Net, on RIGA (see Table~\ref{tab:RIGA-comparision}) on the QUBIQ dataset.
Table~\ref{tab:QUBIQ_comparison} shows the obtained `Average' scores of CM-Net and our PADL. 
It can be seen that the PADL model performed better than CM-Net on all five tasks of the QUBIQ dataset, again demonstrating PADL's ability of modeling annotators' preferences.

\begin{table*}[t]
\small
\renewcommand\arraystretch{1.05}
\setlength\tabcolsep{6.5pt}
\centering
\caption{
Soft Dice ($\mathcal{D}_{avg}^s$(\%), $\mathcal{D}_{mv}^s$(\%)) of PADL and its two variants on five tasks of the QUBIQ dataset.
The predictions are evaluated against each annotator's delineations ($\mathcal{D}_{avg}^s$: Average soft Dice over all annotators) and mean voting annotation ($\mathcal{D}_{mv}^s$).
}
\begin{tabular}{l|c|c|c|c|c}
\hline \hline
Methods & $\mathcal{D}_{kidney}^{s}$ & $\mathcal{D}_{brain}^{s}$ & $\mathcal{D}_{tumor}^{s}$ & $\mathcal{D}_{pros1}^{s}$ & $\mathcal{D}_{pros2}^{s}$ \\ \hline \hline
Ours w/o SEM & (80.24, 79.84) & (80.83, 85.36) & (86.34, 87.63) & (91.39, 93.13) & (74.92, 80.05) \\ \hline
Ours w/o HPM & (79.36, 80.32) & (80.45, 85.73) & (86.80, 88.32) & (90.45, 93.11) & (74.84, 79.53) \\ \hline 
Ours         & (\textbf{82.26}, \textbf{80.34}) & (\textbf{83.19}, \textbf{85.86}) & (\textbf{87.58}, \textbf{89.25}) & (\textbf{92.26}, \textbf{93.30}) & (\textbf{75.16}, \textbf{80.67}) \\ \hline \hline
\end{tabular}
\label{tab:QUBIQ_Ablation}
\end{table*}

\subsection{Ablation Analysis}

Both the SEM and HPM modules play an essential role in the proposed PADL framework, modeling the stochastic errors and annotators' preference independently. We conducted ablation studies on the RIGA dataset to investigate the effectiveness of these two modules, respectively.

\begin{table}[t]
\small
\renewcommand\arraystretch{1.05}
\setlength\tabcolsep{1.5pt}
\centering
\caption{
Soft Dice (\%) of the PADL with complete SEM module and its three variants on the RIGA dataset.
The ground truth is the mean voting annotation, and the `Mean Voting' of each task (optic disc/cup segmentation) is calculated.
}
\begin{tabular}{l|lll|l|l}
\hline \hline
\multicolumn{1}{c|}{\multirow{2}{*}{Baseline}} & \multicolumn{3}{c|}{SEM module}                                                                & \multicolumn{1}{c|}{\multirow{2}{*}{$\mathcal{D}_{disc}^s$ (\%)}} & \multicolumn{1}{c}{\multirow{2}{*}{$\mathcal{D}_{cup}^s$ (\%)}} \\ \cline{2-4}
\multicolumn{1}{c|}{}                          & \multicolumn{1}{p{1cm}<{\centering}|}{$\sigma$}  & \multicolumn{1}{p{1cm}<{\centering}|}{$E_{\mu}$} & \multicolumn{1}{p{1cm}<{\centering}|}{$\mu$ prior} & \multicolumn{1}{c|}{}                     & \multicolumn{1}{c}{}                     \\ \hline \hline
\multicolumn{1}{c|}{$\surd$}                   & \multicolumn{1}{c|}{}          & \multicolumn{1}{l|}{}          &                                & \multicolumn{1}{c|}{96.42}                  & \multicolumn{1}{c}{85.37}                \\ \hline
\multicolumn{1}{c|}{$\surd$}                   & \multicolumn{1}{c|}{$\surd$}   & \multicolumn{1}{l|}{}          &                                & \multicolumn{1}{c|}{97.58}                  & \multicolumn{1}{c}{86.93}                 \\ \hline
\multicolumn{1}{c|}{$\surd$}                   & \multicolumn{1}{c|}{$\surd$}   & \multicolumn{1}{c|}{ $\surd$}  &                                & \multicolumn{1}{c|}{97.65}                  & \multicolumn{1}{c}{87.45}                 \\ \hline
\multicolumn{1}{c|}{$\surd$}                   & \multicolumn{1}{c|}{$\surd$}   & \multicolumn{1}{c|}{$\surd$}   & \multicolumn{1}{c|}{$\surd$}   & \multicolumn{1}{c|}{\textbf{97.65}}                  & \multicolumn{1}{c}{\textbf{87.75}}                     \\ \hline
\hline
\end{tabular}
\label{tab:Ablation-studies}
\end{table}

\subsubsection{Contributions of SEM and HPM}
To evaluate the contributions of SEM and HPM, we compared the proposed PADL framework with its variants that use only one module, $i.e.$, `Ours w/o SEM' and `Ours w/o HPM'.
In `Ours w/o SEM', the SEM module is replaced with a vanilla segmentation head, which directly converts image features to a segmentation map. Accordingly, the $\sigma$ head in each HPM module is also removed.
`Ours w/o HPM' represents the variant that removes all HPM modules. This variant is, of course, not able to model the preference of each annotator.
The performance of our PADL and its two variants was given in Table~\ref{tab:RIGA-comparision}.
It shows that, when the HPM modules were removed, the performance in reconstructing each annotator's delineations drops seriously, especially in optic cup segmentation, $i.e.$, the average $\mathcal{D}_{cup}^{s}$ over six annotations drops remarkably from 84.18\% to 79.24\%.
However, if we evaluated them against the mean voting annotations, the performance only decreases slightly, indicating again that the HPM module is essential for annotators' preference independently.
This is because when HPMs were removed, the SEM can still generate the meta segmentation, but the annotator preference cannot be characterized.
Meanwhile, it also shows that, without the SEM module, the performance of consensus reconstruction drops from 97.65\% to 96.42\% for optic disc segmentation and from 87.75\% to 85.37\% for optic cup segmentation.
When evaluating with each annotator's delineations, the average performance decline in optic cup segmentation can also be observed, $i.e.$, from 84.18\% to 83.65\%.
Note that since each annotator delineates a sample only once, the annotations of each HPM branch are not sufficient to model annotator-specific stochastic error. Therefore, the performance on `Average' drops slightly when removing SEMs.

We validated the contributions of SEM and HPM on five tasks of the QUBIQ dataset as well (see Table~\ref{tab:QUBIQ_Ablation}). These results consistently indicate that HPM modules contribute to the reconstruction of each annotator's segmentation with corresponding preference, and the SEM module can effectively diminish the impact of stochastic errors and produce accurate consensus reconstruction.

\begin{table}[t]
\small
\renewcommand\arraystretch{1.15}
\setlength\tabcolsep{10.5pt}
\centering
\caption{Performance (average $\mathcal{D}_{disc}^{s}$ (\%), $\mathcal{D}_{cup}^{s}$ (\%) obtained by evaluating against six annotations) of the PADL with different designs of preference estimation block $\mathrm{F_{\mu}^{r}}$ or different input of HPM module on the RIGA dataset. The best results in each block are highlighted in \textbf{bold}.}
\begin{tabular}{cc|c}
\hline \hline
\multicolumn{2}{c|}{Models}                                         & Average                                                            \\ \hline \hline
\multicolumn{1}{c|}{\multirow{6}{*}{$\mathrm{F_{\mu}^{r}}$}}        & 1 Layer $1\times1$ Conv & (63.13, 40.30)                           \\ \cline{2-3} 
\multicolumn{1}{c|}{}                                               & 1 Layer $3\times3$ Conv & (96.55, 83.21)                           \\ \cline{2-3} 
\multicolumn{1}{c|}{}                                               & 1 Layer $5\times5$ Conv & (\textbf{96.66}, 82.93)                  \\ \cline{2-3} \cline{2-3}
\multicolumn{1}{c|}{}                                               & 1 Layer $3\times3$ Conv & (96.55, 83.21)                           \\ \cline{2-3}
\multicolumn{1}{c|}{}                                               & 2 Layers $3\times3$ Conv & (96.52, \textbf{84.18})                 \\ \cline{2-3} 
\multicolumn{1}{c|}{}                                               & 3 Layers $3\times3$ Conv & (96.61, 82.45)                          \\ \hline 
\multicolumn{2}{l|}{HPM w/o $f_{img}$}                              & (96.40, 83.39)                                                     \\ \hline
\multicolumn{2}{l|}{HPM w/o $\mu$}                                  & (95.95, 80.22)                                                     \\ \hline
\multicolumn{2}{l|}{HPM}                                            & (\textbf{96.52}, \textbf{84.18})                                   \\ \hline \hline
\end{tabular}
\label{tab:Ablation-hp-capacity}
\end{table}

\subsubsection{Analysis of Stochastic Error Modeling}
\begin{figure}[t]
  \centering
  \includegraphics[width=0.48\textwidth]{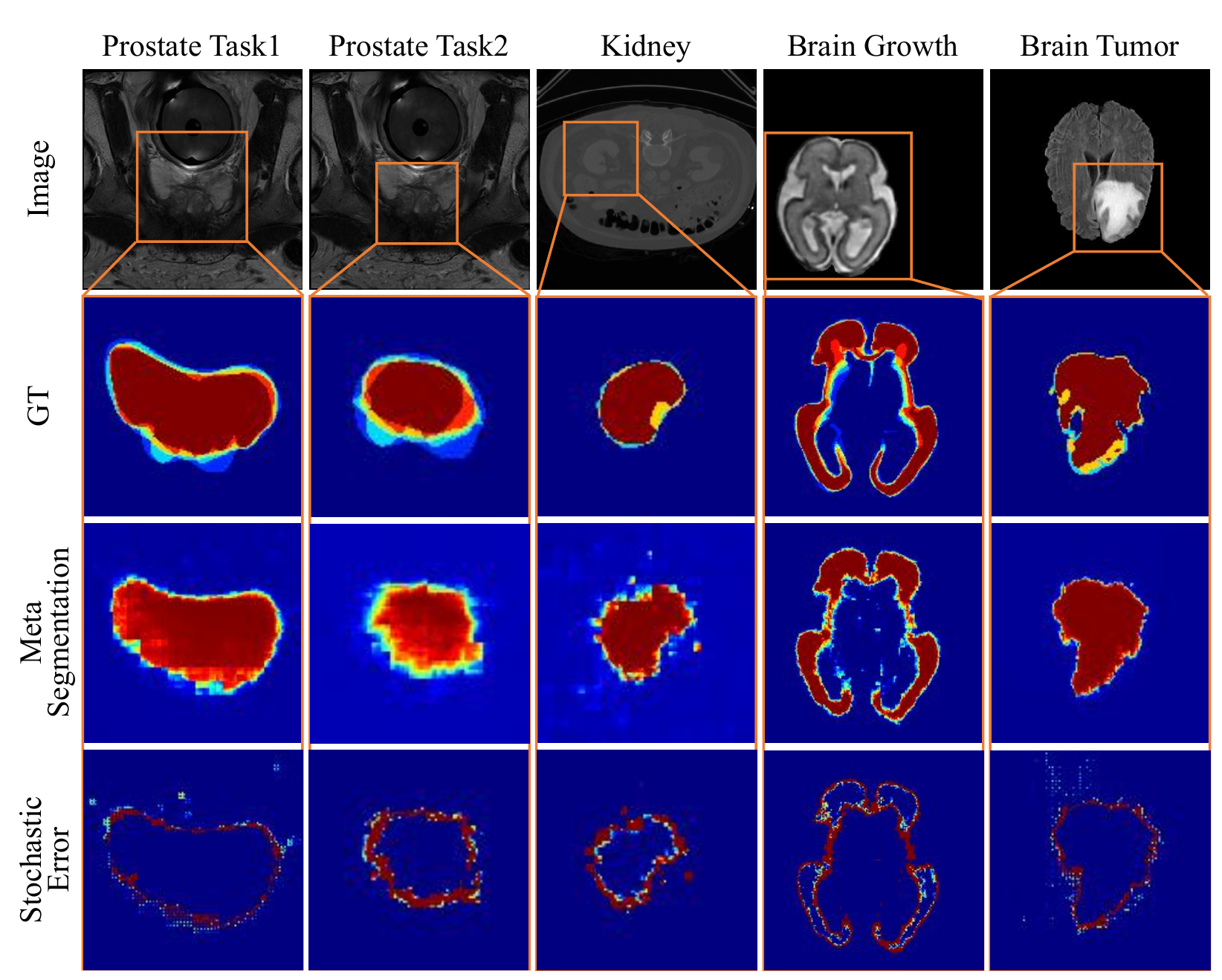}
  \caption{
  Visualization of estimated meta segmentation and stochastic error of five cases from five tasks of the QUBIQ dataset. GT means the mean voting annotation.
  }
  \label{fig:visual_QUBIQ_Ablation}
\end{figure}

\begin{figure}[t]
  \centering
  \includegraphics[width=0.48\textwidth]{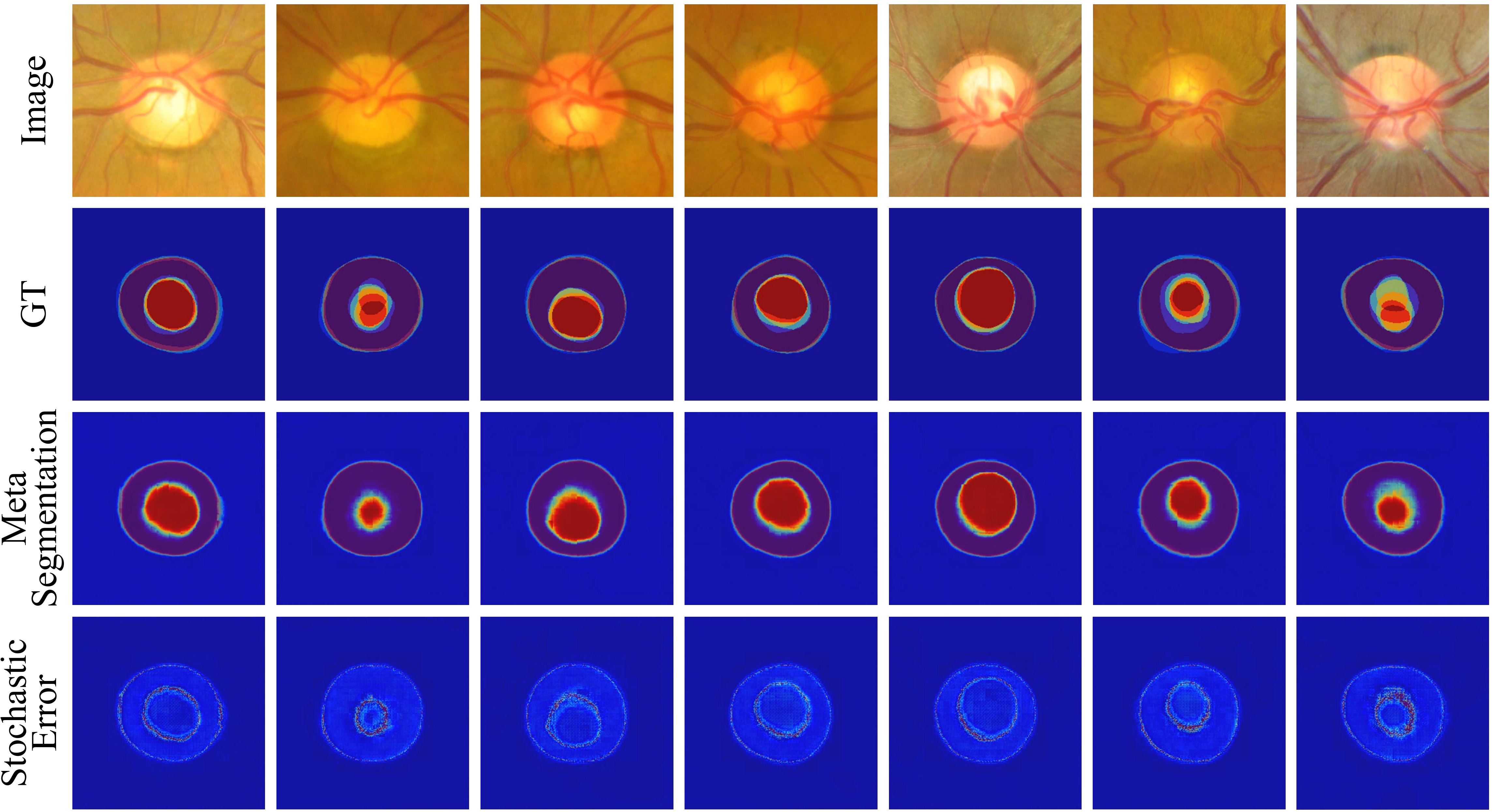}
  \caption{
  Visualization of estimated meta segmentation and stochastic error of seven cases from the RIGA dataset.
  GT means the mean voting annotation.
  }
  \label{fig:Ablation_RIGA}
\end{figure}

The effect of each block in the SEM module was accessed using the mean voting annotations as the ground truth.
Table~\ref{tab:Ablation-studies} gives the performance of the PADL with complete SEM module and its three variants.
It shows that the stochastic error modeling strategy improves $\mathcal{D}_{disc}^s$ and $\mathcal{D}_{cup}^s$ by 0.17\% and 1.72\% respectively, suggesting the necessity of error disentangling from manual annotations.
It also reveals that the entropy attention and the meta segmentation prior can further improve $\mathcal{D}_{cup}^s$ by 0.52\% and 0.3\%, respectively.
It verifies that the proposed EGA block can emphasize the regions where stochastic errors are prone to happen and 
improve the capacity of stochastic error modeling.
We visualized the meta segmentation maps and stochastic error maps obtained by applying PADL to several cases from RIGA dataset and QUBIQ dataset, together with the mean voting annotations (see figure~\ref{fig:Ablation_RIGA} and figure~\ref{fig:visual_QUBIQ_Ablation}).
It shows that our PADL model can characterize the stochastic error effectively and thus produce accurate meta segmentation results, especially in those ambiguous regions which are often accompanied by larger stochastic errors.

\subsubsection{Analysis of Human Preference Modeling}
The HPM module uses the preference estimation block to reconstruct an annotator's preference based on both image features and the meta segmentation.
We attempted different designs of the preference estimation block, ranging from a $1\times1$ convolutional layer to three $3\times3$ convolutional layers, and also tested the HPM module using either image features or meta segmentation as its input.
The average $\mathcal{D}_{disc}^{s}$ and $\mathcal{D}_{cup}^{s}$ evaluated against each annotator's delineations over six annotators were displayed in Table~\ref{tab:Ablation-hp-capacity}.
It shows that the preference estimation block $\mathrm{F_{\mu}^{r}}$ with two of $3\times3$ convolutional layers achieves substantially better $\mathcal{D}_{cup}^{s}$ than that of other settings.
It seems that $1\times1$ kernels cannot model human preference due to its small perception field and $5\times5$ kernels can improve the perception field, but may miss fine details.
Moreover, the performance of the deeper design (with three layers) is limited by the increased optimization difficulty.
Meanwhile, when the meta segmentation $\mu$ is removed from the input, the performance on $\mathcal{D}_{cup}^{s}$ drops remarkably from 84.18\% to 80.22\%, suggesting that the meta segmentation plays a more important role than image features $f_{img}$ in annotator preference estimation.
It can be attributed to the fact that the meta segmentation contains rich-semantic consensus information and is able to enhance the perception of the preference estimation.
Without the mata segmentation, each preference estimation head has to reconstruct each annotator's delineation from scratch based merely on $f_{img}$, which is hard to achieve.

\begin{table*}[t]
\small
\centering
\renewcommand\arraystretch{1.05}
\setlength\tabcolsep{4.5pt}
\caption{
Annotator preference counted on training set (top two rows), test set (middle two rows), and predicted segmentation maps (bottom two rows) on the RIGA dataset. 
The IoU between each annotation/segmentation and the union of six annotations/segmentation maps is utilized to quantify the annotator's preference. 
The number in each bracket is the rank of IoU from highest to lowest in each row. $A_{r}$ means the $r$-th annotator.
}
\begin{tabular}{cc|c|c|c|c|c|c}
\hline \hline
\multicolumn{2}{c|}{(\%)}  & \textcolor[RGB]{26,71,152}{A1}    & \textcolor[RGB]{72,132,32}{A2}    & \textcolor[RGB]{112,112,112}{A3}    & \textcolor[RGB]{209,148,1}{A4}    & \textcolor[RGB]{160,160,160}{A5}    & \textcolor[RGB]{190,79,7}{A6}    \\ \hline \hline
\multicolumn{1}{c|}{\multirow{2}{*}{Average IoU on Training Set}}   & Disc & 89.46 (5) & 88.29 (6) & 93.40 (1) & 90.63 (4) & 92.93 (2) & 90.97 (3) \\ \cline{2-8} 
\multicolumn{1}{c|}{}                       & Cup  & 69.09 (3) & 71.74 (2) & 62.80 (4) & 91.68 (1) & 56.58 (5) & 56.37 (6) \\ \hline
\multicolumn{1}{c|}{\multirow{2}{*}{Average IoU on Test Set}}   & Disc & 88.26 (6) & 88.50 (5) & 94.76 (1) & 91.49 (2) & 92.77 (3) & 89.55 (4) \\ \cline{2-8} 
\multicolumn{1}{c|}{}                       & Cup  & 67.25 (3) & 73.56 (2) & 65.75 (4) & 91.34 (1) & 58.22 (5) & 54.36 (6) \\ \hline
\multicolumn{1}{c|}{\multirow{2}{*}{Average IoU on Segmentation Maps}} & Disc & 96.14 (5) & 95.48 (6) & 99.00 (1) & 96.48 (4) & 98.47 (2) & 96.85 (3) \\ \cline{2-8} 
\multicolumn{1}{c|}{}                       & Cup  & 70.15 (3) & 75.34 (2) & 62.08 (4) & 99.35 (1) & 56.49 (6) & 57.49 (5) \\ \hline
\hline
\end{tabular}
\label{tab:human_preference}
\end{table*}

\begin{figure*}[t]
  \centering
  \includegraphics[width=0.85\textwidth]{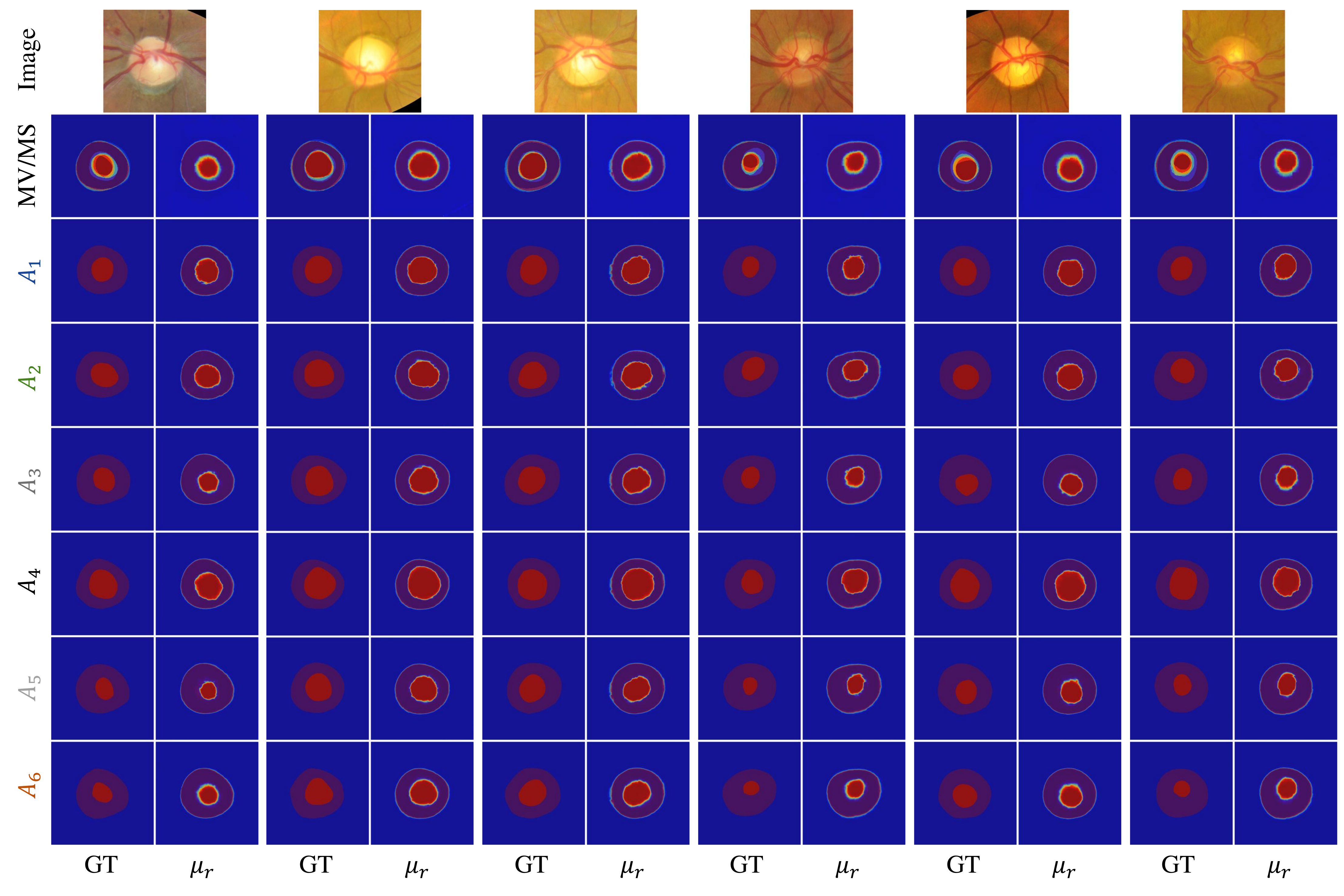}
  \caption{Visualization of probabilistic segmentation maps obtained by applying our PADL modle to six cases from the RIGA dataset, together with ground truths (GTs).
  For each case, the left column shows the Mean Voting GT ($i.e.$, MV) and GTs from six annotators, and the right column shows the meta segmentation ($i.e.$, MS) and six annotator-specific segmentation results $\left \{ \mu_{r} \right \}_{r=1}^{R}$ ($R$=6). $A_{r}$ means the $r$-th annotator. GT ($i.e.$, ground truth) means a manual annotation.
  }
  \label{fig:visual_RIGA_HP_Ablation}
\end{figure*}

For each test sample in the RIGA dataset, the proposed PADL model can generate six segmentation maps, which approximate the annotations given by six annotators, respectively.
We use the IoU between each segmentation and the union of six segmentation maps to quantify the predicted annotator's preference.
To verify whether the predicted annotator's preference is consistent with that embedded in annotations, we calculate the IoU between the annotation of each annotator and the union of six annotations on the training set and test set, too.
The average IoU values counted on the training set, test set and predicted segmentation maps, together with the rank of IoU over six annotators, are give in Table~\ref{tab:human_preference}.
It shows that the rank of predicted annotators' preferences and the rank of annotators' preferences counted over the training/test set are almost identical, indicating that our PADL model is able to characterize each annotator's preference and mimic him/her to perform medical image segmentation.
In figure~\ref{fig:visual_RIGA_HP_Ablation}, we visualize the annotator-specific segmentation results of six cases randomly selected from the RIGA dataset.
It shows that our PADL model can mimic annotator $A_{2}$ and annotator $A_{4}$ to generate larger optic cups, and can mimic annotator $A_{3}$ to generate larger optic discs. Such learned preference is completely consistent with the given annotations.

\section{Conclusion}
In this work, we highlight the issue of annotator-related biases existed in the field of medical image segmentation and propose the PADL framework, which treats the annotation bias as the combination of annotator's preference and stochastic errors, and hence design the SEM module and annotator-specific HPM modules to characterize each annotator's preference while diminishing the impact of stochastic errors.
To our best knowledge, this is the first work that simultaneously explicitly models the annotator preference and disentangles the stochastic annotation error by learning the annotation distribution.
Experimental results on two medical image segmentation benchmarks show that our PADL framework performs well on modeling human preference and disentangling stochastic errors, and achieves better performance against other methods for medical image segmentation with biased annotations.

\section*{Acknowledgments}

We acknowledge the endeavors devoted by the authors of Ref.~\cite{almazroa2017agreement} to annotate and share the fundus imaging data for comparing annotator bias-involved optic cup and optic disc segmentation algorithms.
We also appreciate the efforts devoted by the organizers and sponsors of the Quantification of Uncertainties in Biomedical Image Quantification (QUBIQ) Challenge to collect and share the data for comparing automated medical image segmentation algorithms.

\bibliographystyle{IEEEtran}
\bibliography{reference}

\end{document}